\pdfoutput=1

\documentclass[11pt]{article}

\usepackage{EMNLP2023}

\usepackage{times}
\usepackage{latexsym}

\usepackage[T1]{fontenc}

\usepackage[utf8]{inputenc}

\usepackage{microtype}

\usepackage{inconsolata}
\usepackage{graphicx}
\usepackage{subcaption}
\usepackage{listings}
\usepackage{amsmath}
\usepackage{amssymb}
\usepackage{xcolor}   
\usepackage{multirow}

\colorlet{lightyellow}{yellow!40}

\newcommand\mc{\mathcal}
\newcommand\mb{\mathbf}
\newcommand{\lhp}{LHP }
\newcommand{\lhpns}{LHP}
\newcommand{\lhpfull}{learning from human preferences }

\makeatletter
{\small 
\xdef\f@size@small{\f@size}
\xdef\f@baselineskip@small{\f@baselineskip}
\normalsize 
\xdef\f@size@normalsize{\f@size}
\xdef\f@baselineskip@normalsize{\f@baselineskip}
}
\newcommand{\smalltonormalsize}{%
  \fontsize
    {\fpeval{(\f@size@small+\f@size@normalsize)/2}}
    {\fpeval{(\f@baselineskip@small+\f@baselineskip@normalsize)/2}}%
  \selectfont
}
\makeatother

\title{Pedagogical Alignment of Large Language Models}

\author{
  Shashank Sonkar$^*$\\
  Rice University \\
  \texttt{ss164@rice.edu} \\
  \And
  Kangqi Ni$^*$\\
  Rice University \\
  \texttt{kn22@rice.edu} \\
  \AND 
  Sapana Chaudhary\\
  Texas A\&M University \\
  \texttt{sapanac@tamu.edu} \\
  \And
  Richard G. Baraniuk \\
  Rice University \\
  \texttt{richb@rice.edu} \\
  \\
}

\begin{document}
\maketitle

\begin{abstract}
Large Language Models (LLMs), when used in educational settings without pedagogical fine-tuning, often provide immediate answers rather than guiding students through the problem-solving process. 
This approach falls short of pedagogically best practices and limits their effectiveness as educational tools. We term the objective of training LLMs to emulate effective teaching strategies as `pedagogical alignment.'
In this paper, we investigate Learning from Human Preferences (\lhpns) algorithms to achieve this alignment objective. 
A key challenge in this process is the scarcity of high-quality preference datasets to guide the alignment.
To address this, we propose a novel approach for constructing a large-scale dataset using synthetic data generation techniques, eliminating the need for time-consuming and costly manual annotation.
Leveraging this dataset, our experiments with Llama and Mistral models demonstrate that LHP methods outperform standard supervised fine-tuning (SFT), improving pedagogical alignment accuracy by 13.1\% and 8.7\% respectively.
Existing evaluation methods also lack quantitative metrics to adequately measure the pedagogical alignment of LLMs.
To address this gap, we propose novel perplexity-based metrics that quantify LLMs' tendency to provide scaffolded guidance versus direct answers, offering a robust measure of pedagogical alignment.
Our analysis provides compelling evidence for the superiority of \lhp methods over SFT in optimizing LLMs' behavior, underscoring the potential of \lhp methods in better aligning LLMs with educational objectives and fostering effective learning experiences. 
Code and models are available \href{https://github.com/luffycodes/Tutorbot-Spock}{here}.

\end{abstract}

\renewcommand{\thefootnote}{}
\footnote{$^*$Equal contribution.}

\section{Introduction}
Large Language Models (LLMs) are increasingly being aligned with the `3H' principle - honesty, harmlessness, and helpfulness \citep{askell2021general}.
However, these definitions often require domain-specific adaptations to be truly effective.
In educational contexts, for instance, `helpfulness' takes on a new meaning; rather than providing immediate answers, LLMs should guide students through a structured step-by-step problem-solving processes to encourage active learning and critical thinking \citep{paul1995critical,prince2004does,chi2009active}.
This objective, which we term \textbf{`pedagogical alignment,'} involves breaking complex problems into manageable steps and providing hints and scaffolded guidance rather than direct answers \citep{bransford2000how,hattie2007power}.
Previous adaptation of LLMs for education, such as the CLASS framework \citep{sonkar2023class}, relied on supervised fine-tuning (SFT). However, SFT doesn't explicitly favor pedagogically effective responses, a key aspect of pedagogical alignment \citep{christiano2017deep,ziegler2019fine}.

\begin{figure*}[t]
    \centering
    \includegraphics[width=1.8\columnwidth]{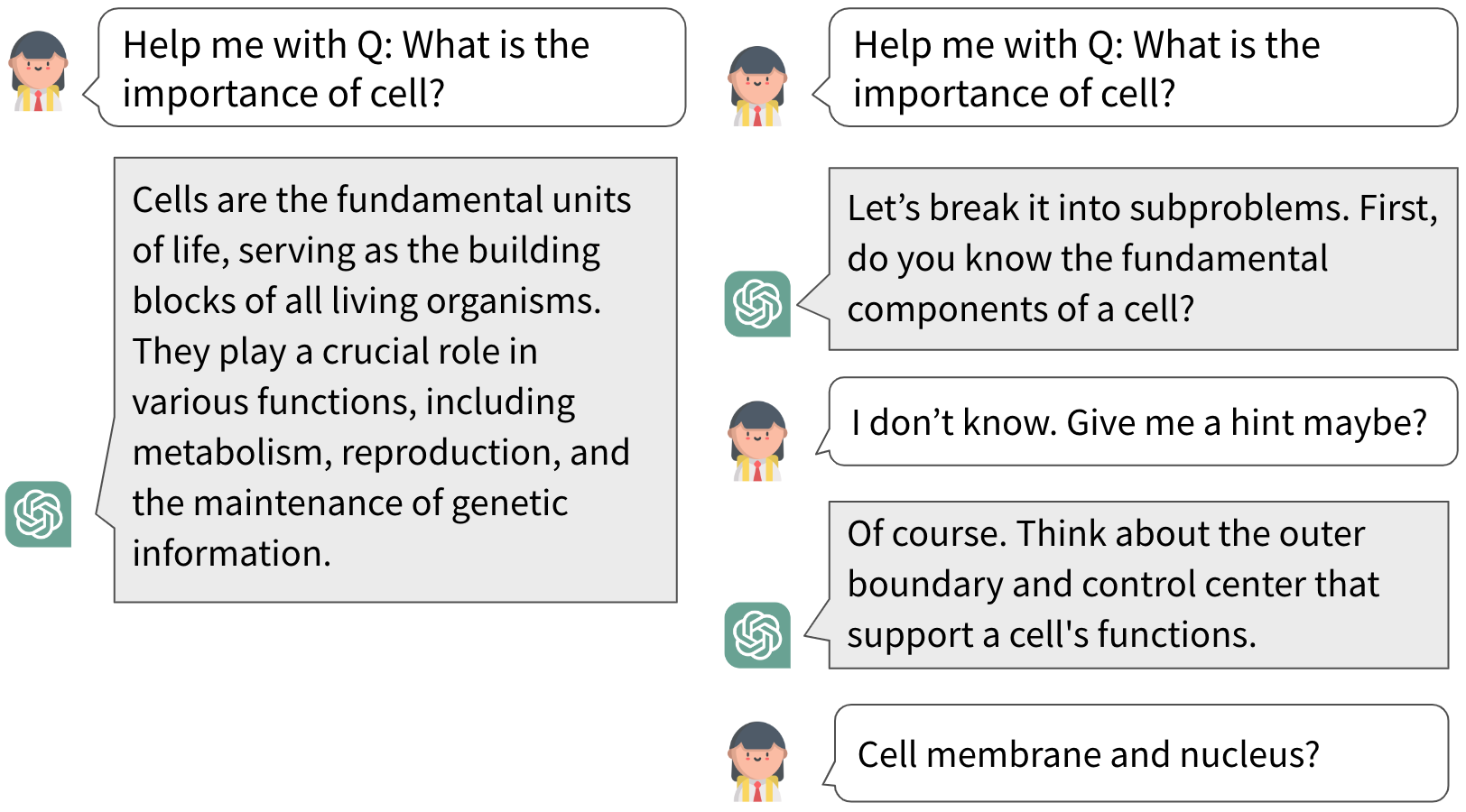}
    \caption{The image depicts the comparison between a traditional Large Language Model (LLM) interaction (left) and a pedagogically-aligned LLM interaction (right). The traditional LLM directly provides the user with the answer, while the pedagogically-aligned LLM guides the student to the solution by presenting a series of subproblems. This elucidates the concept of pedagogical alignment, emphasizing the transformation from direct problem-solving to a guided, scaffolded learning experience.}
    \label{fig:example}
\end{figure*}

In this paper, we propose a novel approach to achieve pedagogical alignment by modeling it as \lhpfull (\lhpns).
This method allows us to represent desired teaching behaviors as preferences, enabling more nuanced optimization than SFT. 
However, a significant challenge in applying LHP is the scarcity of high-quality, pedagogically-sound preference data. To address this, we introduce a key innovation: leveraging the structured output from the CLASS framework \citep{sonkar2023class} to create meaningful preference pairs of pedagogically aligned and misaligned responses. 
This enables us to generate a large-scale, synthetic preference dataset that transforms abstract tutoring principles into concrete, comparative examples, allowing \lhp algorithms to learn effective teaching strategies.

Next, using this preference dataset, we study the effectiveness of three \lhp algorithms: Direct Preference Optimization (DPO) \citep{rafailov2023direct}, Identity Preference Optimization (IPO) \citep{azar2023general}, and Kahneman Tversky Optimization (KTO) \citep{ethayarajh2024kto} for pedagogical alignment of three open-source LLMs - Llama \cite{llama3}, Mistral \citep{jiang2023mistral} and Zephyr \citep{tunstall2023zephyr}.
We evaluate the LLMs using pedagogical alignment accuracy and F1 score which measure if the models produce the desired pedagogical responses. 
Our results show that DPO and KTO significantly improve Llama, Mistral and Zephyr's performance, \textit{improving pedagogical alignment accuracy by 13.1\%, 8.7\% and 50.0\%}, respectively, compared to SFT.

Existing evaluation methods for LLMs in education also fail to adequately quantify pedagogical alignment.
To address this gap and compare \lhp with SFT, we introduce a second preference dataset generation technique and a novel perplexity-based evaluation approach.
This allows us to quantitatively measure an LLM's tendency to provide step-by-step guidance versus direct answers, offering a robust metric for pedagogical alignment.
Using our technique we generate pedagogically misaligned actions, which we then compared to the aligned actions in terms of perplexity. 
Our analysis revealed that base models are more inclined to provide direct solutions than offer hints and guidance. While SFT partially corrects this behavior, it doesn't fully optimize it. 
However, \lhp dramatically optimizes this behavior, promoting guidance over direct solutions. This shift in behavior underscores the effectiveness of \lhp in aligning LLMs with educational objectives, providing compelling evidence for its superiority over SFT in fostering effective learning experiences.

\section{Related Work}
\begin{table*}[t!]
\vspace{0mm}
\centering
\begin{minipage}[t]{0.47\textwidth}
\begin{lstlisting}[mathescape=true,basicstyle=\ttfamily\small]
$\textbf{Prompt}$: Your goal is to create a mock conversation between Student and a Tutorbot, an AI-powered chatbot designed to help Student's with a question:

"Student": "Q. {problem}",
"Evaluation of Student Response": "..",
"Action Based on Evaluation": "..",
"Subproblem State": "..",
"Tutorbot": "Let's break the problem into subproblems and tackle the subproblems one by one. Let's begin with the first subproblem...",

Evaluation of Student Response:
a) Evaluating Incorrect Responses
b) Evaluating Correct Responses
c) Evaluating Partially Correct Responses
d) Evaluating Ambiguous or Unclear or Short Responses
e) Redirecting Off-topic Responses
f) Responding to Student Inquiries
g) N/A

\end{lstlisting}
\end{minipage}
\hfill
\begin{minipage}[t]{0.47\textwidth}
\begin{lstlisting}[mathescape=true,basicstyle=\ttfamily\small]
$\textbf{Prompt continues:}$
If "a" is the evaluation, then:
$\colorbox{lightyellow}{Action 1)}$ Promptly notify the student about the mistake, Provide constructive feedback to pinpoint the errors, Offer helpful hints
$\colorbox{lightyellow}{Action 2)}$ Step in to provide a solution if the student is unable to answer even after multiple attempts.
...
If "c" is the evaluation, then:
$\colorbox{lightyellow}{Action 4)}$ Acknowledge the accurate parts, Promptly notify the student about the mistake, Provide constructive feedback to pinpoint the errors, Offer helpful hints
$\colorbox{lightyellow}{Action 5)}$ Step in to provide a solution if the student is unable to answer after multiple attempts
...

Subproblem states=:
x) One of the subproblems is currently being solved 
y) Subproblem finished, moving to next subproblem
...

\end{lstlisting}
\end{minipage}
\caption{Synthetic conversational data generation prompt of CLASS framework. While CLASS introduced actions for interpretability, in our work we utilize these actions in a novel way to construct a preference dataset that distinguishes pedagogically preferred responses from less effective ones for training our \lhp models.}
\label{tab:prompt_inf}
\end{table*}

\subsection{Algorithms for \lhp}
\label{sec:rlalgos}

Recent developments in the field of \lhpfull (\lhpns) and reinforcement learning through human feedback (RLHF) present promising alternatives for pedagogical alignment.
RLHF refines LLM's behavior based on human-derived preferences or feedback, promoting a more profound congruence with human ethical standards and objectives. 
The seminal work by \citep{ziegler2019fine} introduced the concept of leveraging human preferences to steer the fine-tuning process of language models, thereby laying the foundational stone for RLHF.
Subsequent advancements \citep{nakano2021webgpt,ouyang2022training} have led to notable enhancements in the RLHF pipeline, augmenting alignment efficiency and overall model performance.

More recently, Direct Preference Optimization (DPO) \citep{rafailov2023direct} has emerged as a streamlined and robust advancement over RLHF, offering superior performance by forgoing the necessity for an explicit reward model.
DPO enhances alignment efficiency by directly optimizing a preference-based loss function, thereby simplifying the implementation and the operational efficiency. DPO is designed to directly leverage a dataset of preferences, represented as ${(x,y_w,y_l)}$, where each tuple consists of a prompt $x$, a preferred response $y_w$, and a dis-preferred response $y_l$. Let, $\pi_{\theta}$ be the LLM being finetuned, and $\pi_{\text{ref}}$ be a reference LLM (generally an SFTed model). Then, DPO's optimization problem can be written as: 




\begin{multline}
    W = \beta \log \frac{\pi_\theta(y_w | x)}{\pi_{\text{ref}}(y_w | x)}, 
    L = \beta \log \frac{\pi_\theta(y_l | x)}{\pi_{\text{ref}}(y_l | x)}, \\
    L_{\text{DPO}}(\pi_\theta; \pi_{\text{ref}}) = -\mathbb{E}_{(x,y_w,y_l) \sim D} \left[ W - L \right]
\end{multline} 
\label{eq:dpo}

where $\sigma$ is the sigmoid function, and $\beta$ is an algorithmic parameter. 

DPO tends to overfit to the preference dataset and is sensitive to hyperparameter tuning. Recently Identity Preference Optimization (IPO) \citep{azar2023general} was introduced to address DPO’s overfitting issues.
With addition of a regularization term to the DPO's loss function, IPO also trains models to convergence without requiring early stopping using: 
\begin{multline*}
    L_{\text{IPO}}(\pi_\theta; \pi_{\text{ref}}) =
    -\mathbb{E}_{(x,y_w,y_l) \sim D} \\ \left[ \left( \log \frac{\pi_\theta(y_w | x)}{\pi_{\text{ref}}(y_w | x)} -  \log \frac{\pi_\theta(y_l | x)}{\pi_{\text{ref}}(y_l | x)} - \frac{1}{2\beta} \right) ^2 \right]
\end{multline*}

Both DPO and IPO methodologies necessitate a dataset comprising paired preferences, denoted as ${(x,y_w,y_l)}$. 
The assembly of such datasets in a real-world context is notably labor-intensive and financially burdensome. Kahneman-Tversky Optimization (KTO) \citep{ethayarajh2024kto} presents a solution to this challenge by introducing a loss function formulated exclusively on the basis of singular instances identified as ``good'' or ``bad''.

All three methods, DPO, IPO, and KTO, require a dataset consisting of prompts, accepted responses, and rejected responses.
The curation of such datasets in a real-world context is notably labor-intensive and financially burdensome.
However, in the subsequent section, we introduce our innovative reward/preference data generation approach, which effectively circumvents these challenges, paving the way for cost-effective pedagogical alignment of LLMs.

\subsection{Synthetic Student Data Generation}
\label{sec:peda}
To create our preference dataset, we adapt the CLASS framework \citep{sonkar2023class} that uses GPT-4 \cite{gpt4sparkai} to generate synthetic conversational datasets. These datasets simulate a range of interactions between students and GPT-4 using various prompts. An example prompt is illustrated in Table \ref{tab:prompt_inf}. The prompt instructs the model to first evaluate the student response (Evaluation of Student Response), consider which action to employ (Action based on Evaluation), determine the sub-problem (Subproblem State), and conditioned on these fields generate the reply to the student.
While the CLASS framework introduced these fields to enhance interpretability for teachers and other educational stakeholders, we utilize them differently to construct a preference dataset, as we will discuss in the next section.


\section{Preference Data Generation for Pedagogical Alignment}\label{sec:training_data_generation}
\begin{figure*}[t!]
    \centering
    \includegraphics[width=0.75\textwidth]{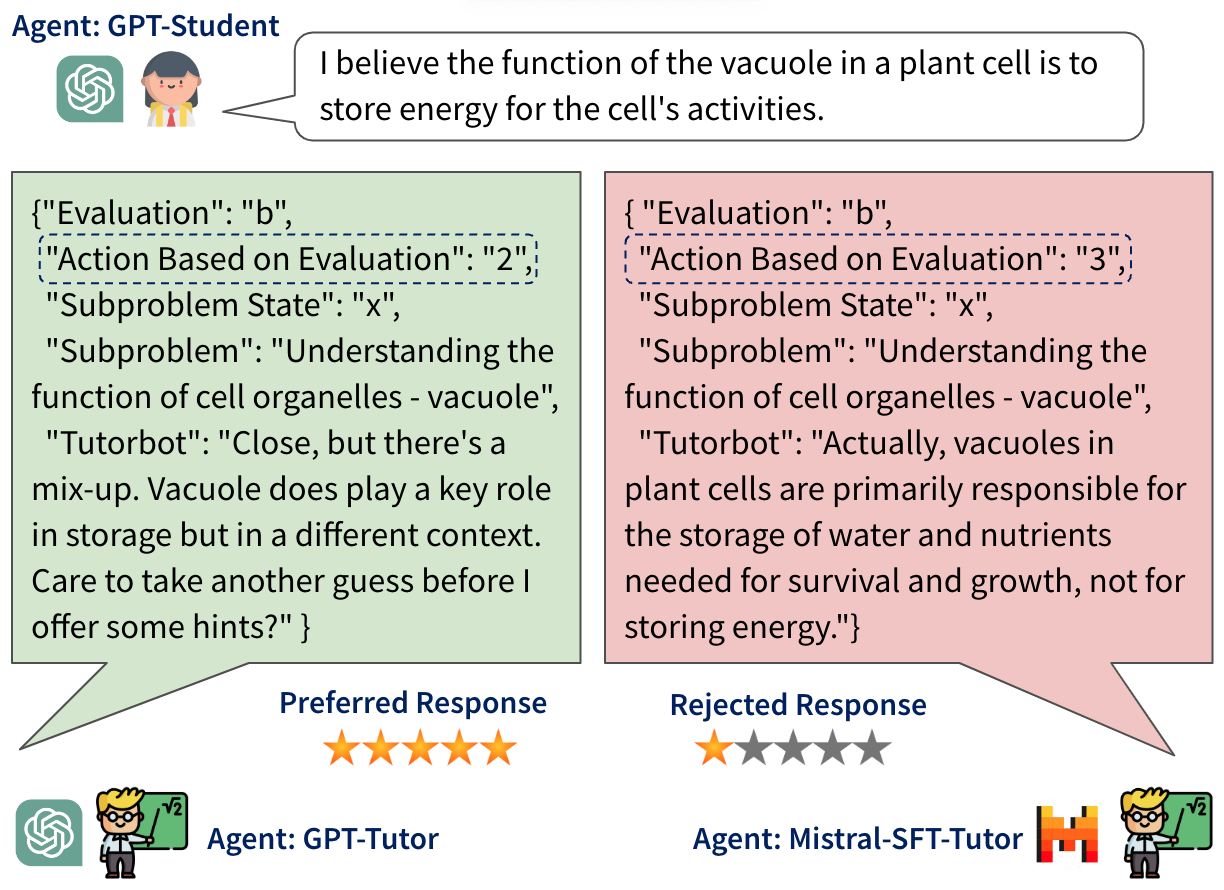}
    \caption{This figure shows the process of generating pedagogically-aligned preference data using the CLASS framework. 
    The GPT-student asks a question, to which both the GPT-tutor and SFT-tutor respond.
    The key focus here is on the divergence in the `Action Based on Evaluation' between the two tutors. 
    In this example, the GPT-tutor's response is deemed more pedagogically aligned because it encourages the student to engage in critical thinking and attempt the problem again, instead of directly providing the correct answer.
    This action mismatch between the two tutor responses allows us to construct a preference dataset that distinguishes between the pedagogically preferred (chosen) and less effective (rejected) responses. 
    }
    \label{fig:rlhf_datagen}
\vspace{-5mm}
\end{figure*}

In this section, we describe our DPO, IPO, and KTO compatible data generation pipeline which we also outline in Figure~\ref{fig:rlhf_datagen}.
The objective is to generate a preference dataset $\mc{D}_p$ which necessitates distinguishing between pedagogically \textit{chosen} and \textit{rejected} tutor responses based on their alignment with desired scaffolding strategies.
$\mc{D}_p$ will mimic a dataset structure that is compatible with DPO, IPO, and KTO to optimally represent these preferences.
This dataset has three fundamental components: \textit{context}, \textit{chosen}, and \textit{rejected}.
In our case, the \textit{context} represents the accumulated conversational history between the student and the tutor leading up to a particular interaction point, providing the necessary backdrop for the tutor's next dialogue turn.
The \textit{chosen} contains the pedagogically preferred or optimal tutor responses to the context, while \textit{rejected} includes those tutor responses considered less effective or misaligned with the scaffolding approach.

\textbf{Conversational Dataset $\mc{D}_c$ and Agents:} We use the conversational prompt listed in Table 1 to generate a dataset represented by $\mc{D}_c = \{(\mb{x}_i, \mb{y}_i) | i = 1, \ldots, N\}$, comprising $N$ student-tutor conversational turns.
Each \( \mathbf{x} \) represents a sequence of student utterances, and each corresponding \( \mathbf{y} \) represents the tutor response.

To generate the preference dataset $\mc{D}_p$ from $\mc{D}_c$, we employ three autonomous agents: GPT-student ($\mc{A}_{\text{G}}$), GPT-tutor ($\mc{A}_{\text{T}}$), and SFT-tutor ($\mc{A}_{\text{S}}$).
The conversation starts with GPT-student $\mc{A}_{\text{G}}$ posing a question ($Q$) sourced from a question bank to which both GPT-tutor $\mc{A}_{\text{T}}$ and SFT-tutor $\mc{A}_{\text{S}}$ provide responses in the following JSON structure:

\begin{equation}
\mb{y}_i = \left\{ 
    \begin{array}{l}
    \parbox{1.0\textwidth}{\normalsize \nonumber
        \text{``Eval of Student Response": ``a, b, c, \ldots, g"}\\
        \text{``Action Based on Eval": ``1, 2, 3, \ldots, 12",}\\
        \text{``Subproblem State": ``w, x, y, z",}\\
        \text{``Subproblem": ``...",}\\
        \text{``Tutorbot": ``..." }
    }
    \end{array}
\right.
\end{equation}

\textbf{Defining Context:} We define the \textit{context} $C_t$ for any given response at time $t$ as all preceding conversational turns up to that point:

\begin{equation}
\nonumber
C_t = \{(\mathbf{x}_i^{(t)}, \mathbf{y}_i^{(t-1)}) | i < t\}, 
\end{equation}

where $\mathbf{x}_i^{(t)}$ denotes the sequence of student response till time $t$, and $\mathbf{y}_i^{(t-1)}$ denotes all tutor responses up to, but not including, the response at time $t$. 
This construction provides a detailed backdrop for each next tutor interaction, feeding into the decision mechanism for \textit{chosen} and \textit{rejected} responses.

\textbf{Preference Dataset $\mc{D}_p$ Construction:} 
Given this understanding of \textit{context}, $\mathcal{D}_p$ is structured to encapsulate interactions where the pedagogically preferred (\textit{chosen}) response diverges from the alternative (\textit{rejected}) based on specific pedagogical criteria. 
This divergence is identified by a function $f$ that assesses the pivotal attributes of each response. Hence, $\mathcal{D}_p$ is defined as:

\begin{align}
\mathcal{D}_p = \big\{ & (C_t, \mathbf{y}_t^\text{Chosen}, \mathbf{y}_t^\text{Rejected}) \, | \, \mathbf{y}_t^\text{Chosen} = \mathbf{y}_t^{\mathcal{A}_T}, \nonumber \\ & \mathbf{y}_t^\text{Rejected} = \mathbf{y}_t^{\mathcal{A}_S}, \text{and } f(\mathbf{y}_t^{\mathcal{A}_T}) \neq f(\mathbf{y}_t^{\mathcal{A}_S}), 
\nonumber \\ & \forall t \in \{1, \ldots, N\} \big\}
\nonumber
\end{align}

where $C_t = x, y_w = y_{choosen}, y_l = y_{rejected}$ from DPO equation~\eqref{eq:dpo} from section~\ref{sec:rlalgos}, and the function $f$ is the key to distinguishing between \textit{chosen} and \textit{rejected} tutor responses.
$f$ indicates a discrepancy between $\mathcal{A}_{\text{T}}$'s and $\mathcal{A}_{\text{S}}$'s responses if any of the fields ``Evaluation", ``Action Based on Evaluation", or ``Subproblem State" diverge between $\mc{A}_{\text{T}}$ and $\mc{A}_{\text{S}}$, $f$ asserts $\mc{A}_{\text{T}}$'s response as preferable, under the assumption that $\mc{A}_{\text{T}}$ offers a more accurate pedagogical model, alongside the rejected response from $\mc{A}_{\text{S}}$. Figure \ref{fig:rlhf_datagen} illustrates this conception with a concrete example: GPT-tutor ($\mc{A}_{\text{T}}$) and SFT-tutor ($\mc{A}_{\text{S}}$) provide different responses since the Action fields are different, so we choose the GPT-tutor response as the accepted one and the SFT-tutor response as the rejected one.
This condition is captured through the function $f$—mapping responses to their pedagogical attributes:


\begin{equation}
{\nonumber
f(\mb{y}_i^{\mc{A}}) = \left(
    \mathrm{Eval}(\mb{y}_i^{\mc{A}}),
    \mathrm{Action}(\mb{y}_i^{\mc{A}}),
    \mathrm{Subprob}(\mb{y}_i^{\mc{A}})
\right)}
\end{equation}

\begin{itemize}
    \item $\text{Eval}(\mathbf{y}_i^{\mathcal{A}})$ extracts insights on the tutor's grading of student's solution or question understanding.
    \item $\text{Action}(\mathbf{y}_i^{\mathcal{A}})$ outlines the tutor's recommended actions based on the evaluation.
    \item $\text{Subprob}(\mathbf{y}_i^{\mathcal{A}})$ provides an understanding of the tutor's perception of the student's current understanding or progress on sub-problems within a larger problem context.
\end{itemize}

The evaluation of these attributes by $f$ facilitates an objective basis to deem one response as \textit{chosen} (preferable) and another as \textit{rejected} (less aligned with desired pedagogical outcomes) through a systematic assessment of their pedagogical value and alignment.


\section{Experiments}

In this section, we provide a comprehensive overview of our dataset construction, experimental design, and the subsequent findings derived from the application of state-of-the-art \lhp algorithms to train pedagogically-aligned LLMs.


\begin{table*}[t!]
\centering
\resizebox{\textwidth}{!}{
\begin{tabular}{|l|c|c|c|c|c|}
\hline
\textbf{Model}   & \textbf{Metric} & \textbf{SFT}             & \textbf{DPO}             & \textbf{IPO}             & \textbf{KTO}              \\ \hline
\textbf{Llama} & Acc            & 64 (62, 61, 70)          & \textbf{77 (74, 74, 84)}         & 74 (71, 70, 81)         & 75 (73, 71, 82)           \\ 
               & F1             & 0.51 (0.54, 0.44, 0.55)    & \textbf{0.57 (0.62, 0.51, 0.59)}  & 0.54 (0.57, 0.47, 0.57)  & 0.54 (0.58, 0.47, 0.58)    \\ \hline
\textbf{Mistral} & Acc            & 65 (61, 61, 72)          & \textbf{74 (71, 69, 80)}         & 71 (67, 67, 79)          & 72 (69, 69, 79)           \\ 
                 & F1             & 0.47 (0.49, 0.41, 0.5)  & \textbf{0.52 (0.54, 0.47, 0.56)}  & 0.5 (0.53, 0.45, 0.54)  & 0.51 (0.54, 0.46, 0.53)   \\ \hline
\textbf{Zephyr}  & Acc            & 23 (22, 21, 26)          & 73 (71, 70, 79)          & 72 (70, 69, 78)          & \textbf{75 (72, 72, 81)}           \\ 
                 & F1             & 0.26 (0.29, 0.23, 0.25)   & 0.55 (0.58, 0.45, 0.62)    & 0.52 (0.57, 0.45, 0.55)  & \textbf{0.56 (0.59, 0.47, 0.62)}   \\ \hline
\end{tabular}
}
\caption{SFT vs DPO, IPO, KTO. 
Each cell displays the average accuracy and F1 score across three classification fields: Evaluation of Student Response, Action Based on the Evaluation, and Subproblem State.
We observe consistent improvements in both accuracy and F1 score for all three alignment algorithms—DPO, KTO, and IPO—compared to their SFT counterparts across the models Llama (Llama-3.1-8B-Instruct), Mistral (Mistral-7B-Instruct-v0.2), and Zephyr (zephyr-7b-beta).
Notably, DPO and KTO consistently outperform IPO for all models.
}

\label{tab:exp1}
\end{table*}

\subsection{Dataset and Evaluation}
Our experimental design requires an extensive conversational dataset to train and test SFT and \lhp models effectively.
The CLASS framework used GPT-4 to generate 610 conversations to train their SFT model.
However, we need to generate a larger volume of conversations to train \lhp models and conduct comprehensive testing.
Following the strategy outlined by CLASS, we set about creating these additional conversations.
We used the CLASS scaffolding prompt to generate the problems, subproblems, and associated hints, which were based on biology topics from OpenStax college textbooks \citep{openstax2018biology}.
These problems seed the context of the conversations generated through GPT-4 in the next step using the CLASS conversational prompt.
In these simulated student-tutor conversations, the tutor uses the subproblems and hints to guide the student toward the final answer.
Through this methodology, we successfully generated an additional $1128$ conversations, resulting in $1738$ conversations.
Each conversation, on average, comprises approximately $8$ rounds and $1311$ words.
This dataset provides a solid platform for training and evaluating SFT and \lhp models.

To partition the training and testing set for SFT and LHP, we randomly sampled our conversational dataset into three partitions: SFT training dataset \(D_{c}\) ($600$ conversations), \lhp training dataset \(D_{p}\) ($600$ conversations), and test dataset \(D_{t}\) ($450$ conversations). 
As each round within the conversation is processed into a separate question-answer (QA) pair, \(D_{c}\) contains $4942$ QA pairs, \(D_{p}\) contains $4921$ QA pairs, and \(D_{t}\) contains $3701$ QA pairs.

For the evaluation, we focus on the three classification fields within the model JSON responses described in Section~\ref{sec:training_data_generation}: Evaluation of Student Response, Action Based on Evaluation, and Subproblem State.
We employ accuracy and F1 score in these three fields to measure the pedagogical alignment of LLMs since these metrics offer a comprehensive view of the model's performance in evaluating the correctness of the responses (Evaluation of Student Response), the appropriateness of the actions (Action Based on Evaluation), and the state of problem-solving (Subproblem State). 
Hence, LLMs that achieve higher accuracy and F1 score in these fields are considered to provide more pedagogically aligned assitance to students.

\subsection{Models and and Training Details} \label{sec:models_training_details}
We experimented on three different LLMs: Llama (Llama-3.1-8B-Instruct), Mistral (Mistral-7B-Instruct-v0.2) and Zephyr (zephyr-7b-beta) with beta of 0.1 for DPO, IPO, and KTO across all models.
Our experiments included training and evaluating SFT and \lhp models. It took around $18$ NVIDIA A6000 GPU hours for one cycle (SFT + Data Generation + \lhp + Evaluation). A total of $9$ cycles, which are $162$ GPU hours, are needed to complete all the experiments.
We conducted the SFT experiments using FastChat \citep{zheng2023judging}. 
For the choice of SFT hyperparameters, we refered to parameters used to instruct fine-tune Vicuna \citep{zheng2023judging}: We employed a learning rate of $2 \times 10^{-5}$ with the AdamW optimizer \citep{loshchilov2019decoupled}, a batch size of $16$, a cosine scheduler, a weight decay of $0.05$, a warmup ratio of $0.1$, and a total of 3 epochs.
We conducted the \lhp experiments using TRL \citep{vonwerra2022trl}. 
For the choice of \lhp hyperparamters, 
we use beta of $0.1$, a learning rate of $1e-7$ with the AdamW optimizer, a batch size of $16$, a cosine scheduler, a weight decay of $0.05$, a warmup ratio of $0.1$, and a total of $3$ epochs.

\subsection{Main Findings: SFT vs \lhpns}
\label{exp1}

In this experiment, we compared three alignment algorithms: Direct Preference Optimization (DPO), Identity Preference Optimization (IPO), and Kahneman Tversky Optimization (KTO) on the SFT models. 
We measured the accuracy and F1 score across three classification fields in the responses: Evaluation of Student Response, Action based on Evaluation, and Subproblem State.

We observed substantial performance improvements across all three SFT models—Llama, Mistral, and Zephyr. Specifically, Llama achieved accuracy gains of 13.1\%, 10.1\%, and 10.9\% from DPO, IPO, and KTO, respectively. Mistral demonstrated increases of 8.7\%, 6.1\%, and 7.4\%, while Zephyr showed improvements of 50.0\%, 50.2\%, and 52.0\% (Table \ref{tab:exp1}). A similar trend was observed in F1 scores: Llama exhibited increases of 6.3\%, 2.7\%, and 3.2\%, Mistral showed gains of 5.6\%, 3.7\%, and 4.3\%, and Zephyr saw improvements of 29.3\%, 26.7\%, and 30.2\% for DPO, IPO, and KTO, respectively. Notably, DPO and KTO significantly outperformed IPO across all models. Also note these results are for a fixed beta of $0.1$. As we show in section~\ref{sec:beta}, beta hyperparameter tuning can further boost the performance of \lhp algorithms.


This experiment shows the pivotal role of alignment algorithms in improving the accuracy and F1 score of SFT models. All models -- Llama, Mistral and Zephyr exhibit pronounced improvements using \lhp algorithms compared to SFT.

\subsection{Pedagogical Shifts: Perplexity Comparison of SFT and \lhpns}
For understanding the effectiveness of \lhp algorithms compared to SFT, we propose a second preference dataset generation technique. This technique specifically allows us to perform a perplexity analysis to compare the LLM's tendency to provide hints and guidance versus direct responses.

\textbf{Dataset Construction:} The essence of pedagogical alignment is the LLM's propensity to offer guidance (as seen in Actions 1 and 4) over direct solutions (as in Actions 2 and 5). \textit{By this definition, Action 2 can never precede Action 1, and likewise, Action 5 can never precede Action 4} (for more details about actions please refer to Table~\ref{tab:prompt_inf}).
With this critical insight, we use GPT-4 to generate pedagogically misaligned Actions (Action 2 for the first occurrence of Action 1 and Action 5 for the first occurrence of Action 4). This approach allows us to compare the perplexity of responses corresponding to Action 1 and its misaligned counterpart Action 2, and similarly for Actions 4 and 5.


We then use this newly constructed Action dataset to evaluate the perplexity of responses generated by the original instruct model (Base) to its SFT and \lhp variants. 

\begin{figure*}[t]
    \centering
    \begin{subfigure}[t]{0.34\textwidth}
        \centering
        \includegraphics[height=4cm]{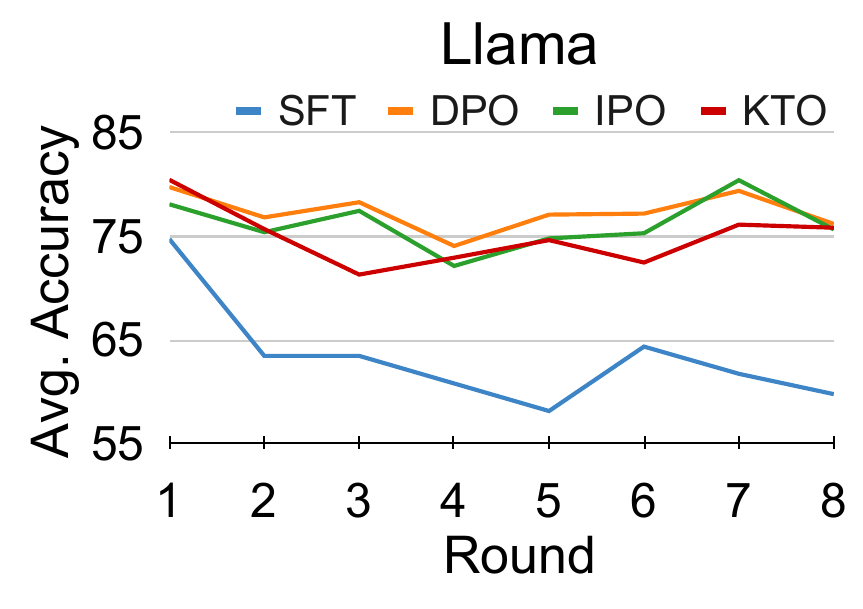}
    \end{subfigure}\hspace{0.01\textwidth}%
    \begin{subfigure}[t]{0.32\textwidth}
        \centering
        \includegraphics[height=4cm]{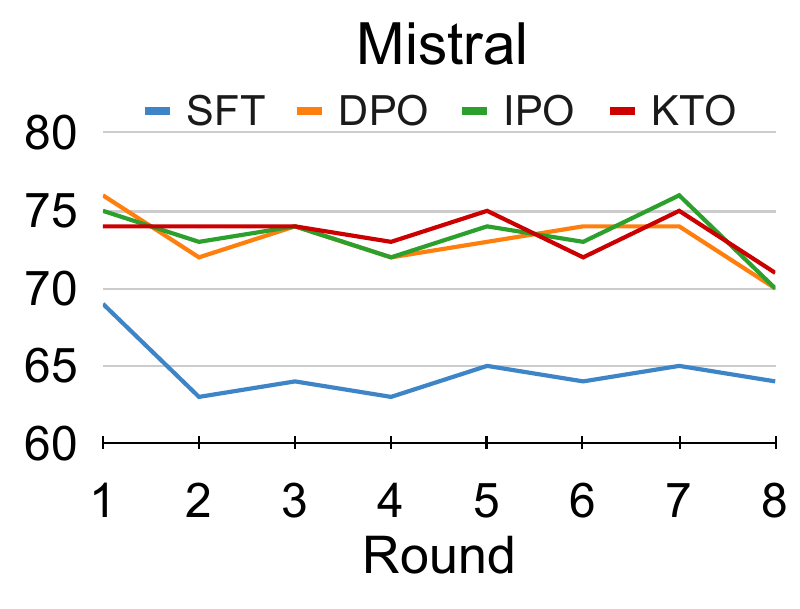}
    \end{subfigure}\hspace{0.01\textwidth}%
    \begin{subfigure}[t]{0.32\textwidth}
        \centering
        \includegraphics[height=4cm]{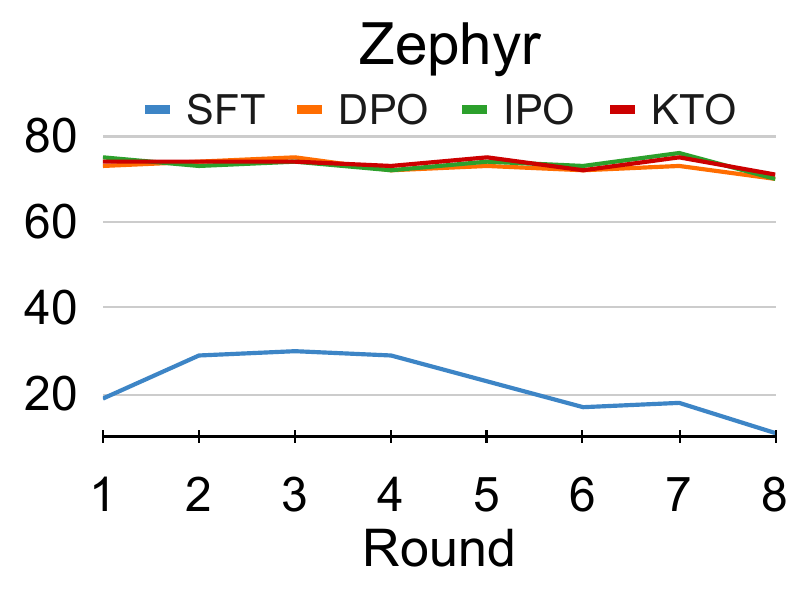}
    \end{subfigure}
    \caption{ Comparison of multi-round performance for SFT vs. \lhp methods (DPO, IPO, KTO) across Llama, Mistral, and Zephyr models. The graphs illustrate average accuracy over 8 conversation rounds, revealing the superior performance of \lhp methods in maintaining pedagogical alignment across extended conversation context.}
    \label{fig:multi-round}
\end{figure*}


\begin{figure*}[t!]
    \centering
    \begin{subfigure}[t]{0.34\textwidth}
        \centering
        \includegraphics[height=4cm]{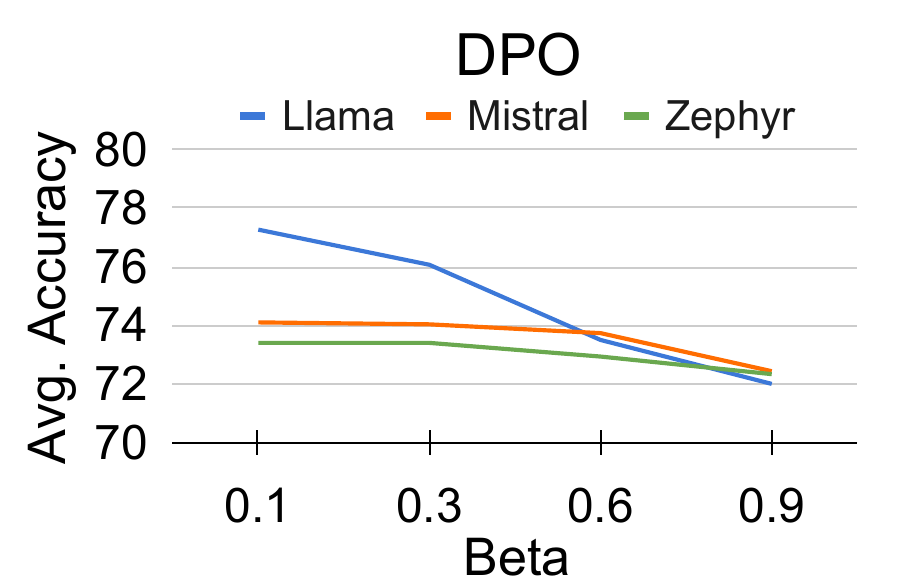}
    \end{subfigure}\hspace{0.01\textwidth}%
    \begin{subfigure}[t]{0.32\textwidth}
        \centering
        \includegraphics[height=4cm]{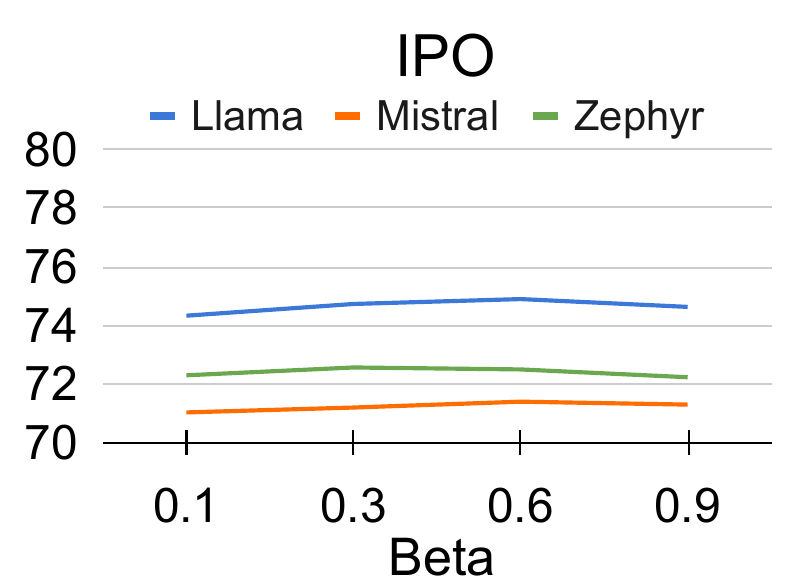}
    \end{subfigure}\hspace{0.01\textwidth}%
    \begin{subfigure}[t]{0.31\textwidth}
        \centering
        \includegraphics[height=4cm]{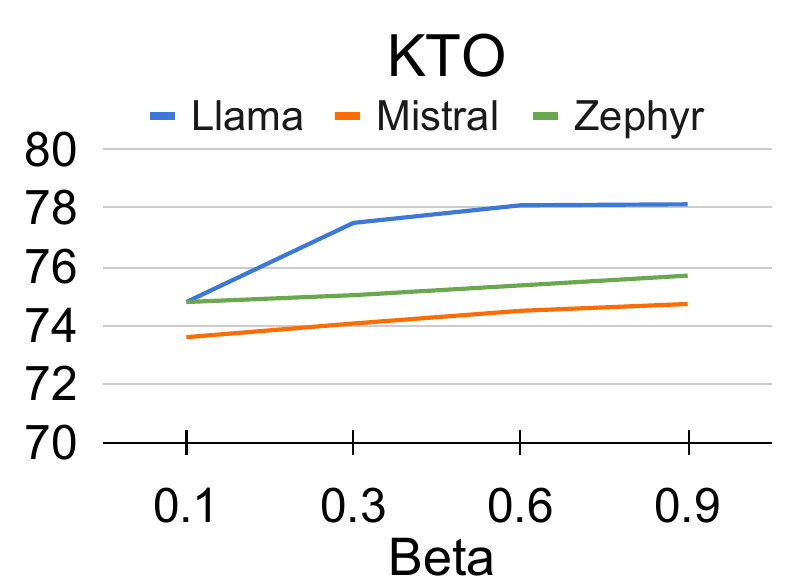}
    \end{subfigure}
    
    \caption{Performance of \lhp algorithms (DPO, IPO, and KTO) as a function of beta. Our results indicate that KTO outperforms both DPO and IPO with optimal beta hyperparameter search.}
    \label{fig:exp2}
    \vspace{-3mm}
\end{figure*}

\textbf{Perplexity Findings:}  Our analysis, as shown in table \ref{tab:ppl}), reveals interesting patterns across the three models:
(A) Llama: The base model shows similar perplexity across all actions, indicating no strong preference for guidance or direct solutions. SFT increases perplexity for Actions 2 and 5, shifting towards more guidance. Both DPO and KTO further optimize this behavior, with DPO showing the lowest perplexity for Actions 1 and 4; (B) Mistral: The base model shows slightly higher perplexity for Actions 1 and 2. SFT dramatically increases perplexity for Actions 2 and 5, strongly favoring guidance. DPO and KTO further reduce perplexity for Actions 1 and 4, with DPO showing slightly better optimization; (C) Zephyr: The base model shows relatively uniform perplexity across actions. SFT significantly increases perplexity for Actions 2 and 5. DPO and KTO further optimize this behavior, with DPO showing lower perplexity for Actions 1 and 4, while KTO shows higher perplexity for Actions 2 and 5.

These results demonstrate that SFT partially shifts model behavior towards providing more guidance, but \lhp methods (DPO and KTO) further optimize this behavior. For Mistral and Zephyr, \lhp methods dramatically increase the perplexity gap between guidance actions (1 and 4) and direct solution actions (2 and 5), indicating a significant shift towards pedagogical alignment.
The perplexity analysis provides compelling evidence for the effectiveness of \lhp methods in achieving pedagogical alignment, consistently outperforming SFT across all three models. 

\begin{table}[h!]
    \centering
    \resizebox{\columnwidth}{!}{
        \begin{tabular}{|l||l|l||l|l||}
        \hline
             \textbf{Model} & \textbf{A1} & \textbf{A2} & \textbf{A4} & \textbf{A5} \\
        \hline
             Llama (Base) & 2.02 & 2.04 & 2.07 & 2.01 \\
             Llama (SFT) & 1.77 & 2.44 & 1.87 & 2.45 \\
             Llama (DPO) & 1.70 & 2.45 & 1.78 & 2.43\\
             Llama (KTO) & 1.74 & 2.48 & 1.81 & 2.46 \\
        \hline
             Mistral (Base)  & 2.11 & 2.06 & 2.01 & 1.97\\
             Mistral (SFT)  & 2.06 & 5.51 & 2.14 & 5.37\\
             Mistral (DPO)  & 1.86 & 5.49 & 1.93 & 5.31\\
             Mistral (KTO) & 1.90 & 5.01 & 1.98 & 4.93 \\
        \hline
             Zephyr (Base) & 2.05 & 2.09 & 2.04 & 2.07\\
             Zephyr (SFT) & 2.09 & 13.9 & 2.18 & 13.06\\
             Zephyr (DPO) & 1.91 & 16.44 & 1.97 & 15.66 \\
             Zephyr (KTO) & 1.97 & 16.76 & 2.03 & 16.16\\
        \hline

        \end{tabular}
    }
    \caption{Perplexity analysis of model responses corresponding to different actions. Lower perplexity for Actions 1 and 4 indicates a higher likelihood of providing scaffolded guidance like hints, while higher perplexity for Actions 2 and 5 suggests a lower tendency to offer direct solutions. Results show SFT partially improves pedagogical alignment, while LHP methods (DPO, KTO) further optimize it by widening the perplexity gap between guidance (Actions 1, 4) and direct solution (Actions 2, 5) responses. For more details about actions please refer to Table~\ref{tab:prompt_inf}.}
    \label{tab:ppl}
\end{table}

\subsection{Pedagogical Consistency Over Time: SFT vs. \lhp in Extended Conversations}

To assess the robustness and consistency of pedagogical alignment across extended conversations, we conducted an in-depth analysis of model accuracy across multiple conversation rounds. This analysis is important because as conversations progress, the context becomes more complex and lengthy, potentially challenging the model's ability to maintain consistent pedagogical alignment.

The analysis reveals striking differences in pedagogical alignment between SFT and \lhp methods across all three models. As can be seen in Figure~\ref{fig:multi-round}, Llama's SFT model shows a clear decline in alignment as conversations progress, dropping from $75\%$ to $60\%$ by round $8$, while \lhp methods maintain consistent alignment around $75\%$. Mistral and Zephyr demonstrate even more pronounced contrasts. Their SFT models struggle with pedagogical alignment in extended conversations, with Mistral fluctuating between $63-69\%$ and Zephyr performing poorly at $11-30\%$. In stark contrast, all \lhp approaches (DPO, IPO, KTO) for both models maintain high pedagogical alignment above $70\%$ throughout, with Zephyr showing the most dramatic improvement over its SFT counterpart. This consistent performance of \lhp methods across all models, particularly in later conversation rounds, underscores their effectiveness in maintaining pedagogical alignment even as context complexity increases.

These results demonstrate that \lhpns-aligned models maintain high pedagogical alignment even with increasing context length, unlike SFT models which tend to degrade in performance over extended conversations. This robustness is crucial for effective tutoring, as it ensures the model can provide reliable guidance throughout a problem-solving session, regardless of its context length.

\subsection{Effect of Beta on \lhp Algorithms}\label{exp2}
\label{sec:beta}

In this experiment, we analyze the impact of the beta parameter on the performance of \lhp algorithms. Beta is a key hyperparameter in all three alignment algorithms (DPO, IPO, and KTO). It controls the strength of the Kullback-Leibler divergence penalty between the trained model and the reference model within the loss function. The choice of beta depends on the specific model and dataset and thus necessitates an empirical study. As we show next, beta hyperparameter tuning can significantly affect the performance of \lhp algorithms, with \textbf{KTO emerging as the top performer}. Similar to experiments in Section \ref{exp1}, we measure the average accuracy and F1 score across the three classification fields. We find that the performance of all three algorithms is sensitive to the choice of beta, but to varying degrees and with different optimal values for each algorithm and model.

For DPO, we observe that lower beta values (0.1 or 0.3) generally yield the best results. Llama achieved the highest average accuracy of 77.3\% and average F1 score of 0.57 with beta as 0.1. Mistral performed best with beta = 0.1, reaching 74.1\% accuracy and 0.52 F1 score. Zephyr showed similar performance for beta = 0.1 and 0.3, with a slight edge for 0.1 (73.4\% accuracy, 0.55 F1 score).

IPO demonstrated less sensitivity to beta changes, especially for Llama and Mistral. Llama's performance peaked at beta = 0.6 with 74.9\% accuracy and 0.54 F1 score. Mistral showed minimal variation, with slightly better results at beta = 0.6 (71.4\% accuracy, 0.51 F1 score). Zephyr performed best with beta = 0.3, achieving 72.6\% accuracy and 0.55 F1 score.

KTO emerged as the top performer, showing consistent improvement with increasing beta values across all models. For Llama, performance peaked at beta = 0.9 with 78.1\% accuracy and 0.57 F1 score. Mistral also showed best results at beta = 0.9, reaching 74.7\% accuracy and 0.51 F1 score. Zephyr demonstrated steady improvement, with optimal performance at beta = 0.9 (75.7\% accuracy, 0.55 F1 score).


Notably, with optimal beta values, KTO outperforms both DPO and IPO across all models, emphasizing its effectiveness in pedagogical alignment tasks when properly tuned. Our findings underscore the necessity of careful hyperparameter tuning in \lhp algorithms to achieve optimal pedagogical alignment. They also suggest that KTO, when properly tuned, may be particularly well-suited for tasks requiring sustained pedagogical alignment.

\section{Conclusion}
In this paper, we have investigated the application of \lhpfull (\lhpns) methods to align LLMs with educational goals, aiming to foster optimal student learning outcomes. By constructing a preference dataset specifically designed for pedagogical alignment, we have laid the foundation for enhancing the effectiveness of LLMs in educational contexts. Our experimental results, derived from applying \lhp alignment algorithms on state-of-the-art open-source LLMs, demonstrate the superiority of \lhp methods over standard supervised fine-tuning (SFT), significantly improving the alignment of LLMs with pedagogical objectives.
Our another key contribution is the development of a novel approach for evaluating the pedagogical effectiveness of LLMs. By curating a preference dataset that compares the perplexity of responses offering scaffolding versus those providing direct solutions, we have introduced a new methodology that quantifies the extent to which LLMs prioritize pedagogically effective actions. This approach opens up new possibilities for assessing and optimizing the performance of LLMs in educational settings, providing a valuable tool for researchers and practitioners alike.

\section{Limitations}
While our methodology shows promising results, there are certain inherent limitations that are challenging to overcome. One such limitation is the dynamic and complex nature of the educational landscape. The effectiveness of our approach may vary depending on a multitude of factors, including the diversity of student learning styles, the evolving nature of educational content, and the rapid advancements in AI technology. Additionally, computational constraints may also limit the scalability of our approach in larger, more diverse educational settings.
More comprehensive user studies can assess the impact of these aligned models in real-world educational settings.

\section{Ethics and Risks}
The use of LLMs in education comes with ethical considerations and potential risks. These include the need to ensure the models do not amplify biases, the importance of protecting sensitive student data, and the necessity of maintaining human guidance in the learning process. In addition, the impact of LLMs on learning outcomes should be continuously evaluated to ensure their effectiveness and avoid unintended negative consequences.

\section*{Acknowledgments}
This work was supported by NSF grant 1842378, ONR grant N0014-20-1-2534, AFOSR grant FA9550-22-1-0060, a Vannevar Bush Faculty Fellowship, and ONR grant N00014-18-1-2047.

\bibliography{anthology}
\bibliographystyle{acl_natbib}



\end{document}